\documentclass[runningheads]{llncs}
\usepackage[T1]{fontenc}
\usepackage{orcidlink}
\usepackage{amsmath,amssymb,amsfonts}
\usepackage{amsfonts}
\usepackage{booktabs}
\usepackage{algorithm}
\usepackage{algorithmic}
\usepackage{siunitx}

\usepackage{comment}

\usepackage{subcaption}

\usepackage{graphicx}
\begin{document}
\title{Combining Facial Videos and Biosignals\\ for Stress Estimation During Driving}

\author{
Paraskevi Valergaki\inst{2}\orcidlink{0009-0005-0216-1267} \and
Vassilis C.~Nicodemou\inst{2}\orcidlink{0000-0001-8523-6749} \and
Iason Oikonomidis\inst{2}\orcidlink{0000-0002-9503-3723} \and
Antonis Argyros\inst{1,2}\orcidlink{0000-0001-8230-3192} \and
Anastasios Roussos\inst{2}\orcidlink{0000-0001-6015-3357} 
}

\authorrunning{Valergaki et al.}

\institute{
Computer Science Department, University of Crete, Heraklion, Greece
\and
Institute of Computer Science (ICS), 
Foundation for Research \& Technology – Hellas (FORTH), Heraklion, Greece\\
\email{\{vbalerg,nikodim,oikonom,argyros,troussos\}@ics.forth.gr}
}

\sloppy

\maketitle              
\begin{abstract}
Reliable stress recognition is critical in applications such as medical monitoring and safety-critical systems, including real-world driving. While stress is commonly detected using physiological signals such as perinasal perspiration and heart rate, facial activity provides complementary cues that can be captured unobtrusively from video. We propose a multimodal stress estimation framework that combines facial videos and physiological signals, remaining effective even when biosignal acquisition is challenging. Facial behavior is represented using a dense 3D Morphable Model, yielding a 56-dimensional descriptor that captures subtle expression and head-pose dynamics over time. To investigate the correlation between stress and facial motions, we perform extensive experiments involving also physiological markers. Paired hypothesis tests between baseline and stressor phases show that 38 of 56 facial components exhibit consistent, phase-specific stress responses comparable to physiological markers. Building on these findings, we introduce a Transformer-based temporal modeling framework and evaluate unimodal, early-fusion, and cross-modal attention strategies. Combining 3D-derived facial features with physiological signals via cross-modal attention substantially improves performance over physiological signals alone, increasing AUROC from $52.7\%$ and accuracy from $51.0\%$ to $92.0\%$ and $86.7\%$, respectively. Although evaluated on driving data, the proposed framework and protocol may readily generalize to other stress estimation settings. Code and additional information is available on the project website - \url{https://vivianval.github.io/stress_driving.html}.


\end{abstract}
\section{Introduction}

Stress recognition from facial videos has gained increasing attention in areas such as affective computing~\cite{ecsa-7-08227}, healthcare~\cite{healthcare}, and intelligent transportation systems~\cite{Siam2023AutomaticSD}, with driver stress monitoring being particularly relevant for road safety. Despite recent advances, reliable video-based stress estimation remains challenging due to the complex and subjective nature of stress, individual variability, and the partially voluntary control of facial expressions, which can be consciously masked or suppressed~\cite{Giannakakis2019ReviewOP}.

Most existing approaches rely either on physiological signals~\cite{PHYSIOLOGICAL} or low-level geometric or appearance features as well as specific coding systems such as Facial Action Units (AUs)~\cite{rel22}, \cite{rel28} which explicitly encode the activation of individual facial muscles. While effective, these representations often entangle identity, pose, and expression, limiting their ability to capture subtle 3D facial dynamics related to stress. Fewer works have explored deep 3D geometric facial features, despite evidence that head pose and fine-grained expression dynamics are sensitive indicators of stress.

In this work, we investigate stress under driving conditions by explicitly modeling the complementary interaction between facial dynamics and physiological signals. We leverage disentangled 3D facial expression and pose parameters extracted from infrared facial videos using EMOCA~\cite{EMOCA:CVPR:2021} and integrate them with physiological measurements (heart rate, breathing rate, and perinasal perspiration) within a unified multimodal framework to achieve robust and efficient stress characterization.

We first perform a statistical analysis using paired tests between baseline and stressor phases to examine stress-related modulations in 3D facial dynamics and physiological signals. Building on these findings, we introduce a Transformer-based temporal modeling framework and systematically evaluate unimodal and multimodal fusion strategies. Our results show that cross-modal attention fusion most effectively captures the interaction between facial and physiological responses, yielding consistent performance improvements of up to $4\%$ AUROC over established machine learning and deep learning baselines.

Our main contributions are threefold: (i)~we demonstrate that fusing dense 3D facial motion features with physiological features significantly improves stress prediction, confirming their complementary role; (ii)~we propose a Transformer-based cross-modal attention framework for effective multimodal fusion that remains robust when only facial videos are available; and (iii)~we focus on driving scenarios, where body cues are limited, highlighting facial motion as a primary source of stress-related visual information. While evaluated on driving data, the proposed methodology may readily extend to other settings involving stress recognition from facial videos.

\section{Related Work}

\noindent\textbf{Stress Recognition Datasets.}
A variety of datasets have been proposed for stress recognition, differing in sensing modalities, elicitation protocols, and annotation strategies. Early datasets such as SUS~\cite{sus} focus on unimodal audio recordings, while video-only datasets such as SADVAW~\cite{sadvaw} derive stress labels from externally annotated movie clips. Several datasets emphasize physiological sensing, including WeSAD~\cite{wesad} and CLAS~\cite{clas}, which collect ECG, EDA, EMG, respiration, and acceleration signals under stress-inducing protocols. More recent multimodal datasets combine physiological measurements with audio and video data, including MuSE~\cite{muse} and SWELL-KW~\cite{swellkw}, though these are limited in scale. Larger datasets such as UBFC-Phys~\cite{UBCuphys}, Distracted Driving~\cite{Distrdriving}, and StressID~\cite{NEURIPS2023_5f09bfe6} provide synchronized facial video and physiological signals, with the latter also including speech recordings.

\noindent\textbf{Stress Detection from Facial Videos.}
Facial activity has been widely studied as a visual indicator of stress, with many approaches relying on Facial Action Units (AUs) and expression-related cues using machine learning and deep learning models~\cite{rel22,rel23,rel24,rel25,rel26,rel27,rel28,rel29,rel30,rel31,rel32,rel33}. Deep AU-based pipelines~\cite{rel27} and attention-driven architectures such as TSDNet~\cite{rel24} report strong stress recognition performance, while regression-based models further demonstrate correlations between facial behavior and perceived stress~\cite{rel32}. Zhang et al.~\cite{rel24} achieved 78.62\% accuracy for face-only stress detection, with further gains (85.42\% accuracy) when incorporating action motion cues. MTASR~\cite{MTASR} extracts rPPG signals from RGB videos and applies multi-task attentional learning, achieving high accuracy on UBFC-Phys. Other studies highlight the role of 3D facial geometry and head pose dynamics in stress and affect analysis~\cite{rel30,Giannakakis2018EvaluationOH}. In contrast to AU-based or shallow geometric approaches, our work employs disentangled 3D facial expression and pose representations and models their temporal dynamics using Transformer-based architectures.

\noindent\textbf{Driving Behavior Classification.}
Driving-related stress has been shown to induce measurable physiological responses~\cite{Huynhetal,HasanCardiovascular2023} and to depend strongly on driving context and visual road conditions~\cite{TAVAKOLI2023101649,Bustos}. Prior work demonstrates that multimodal fusion of visual cues and physiological signals improves driver stress detection~\cite{COMPLEMENTARY1,COMPLEMENTARY2,COMPLEMENTARY3}. Existing approaches integrate eye-tracking and biosignals using classical machine learning~\cite{COMPLEMENTARY1} or employ decision-level fusion with Transformer encoders~\cite{COMPLEMENTARY2}. Other driving behavior classification methods fuse vehicle dynamics with visual data~\cite{FMDNet,GLMDriveNet,MOU2023121066}, without modeling facial behavior. Conversely, our approach adopts bidirectional cross-modal attention to enable direct temporal interaction between facial dynamics and physiological features, allowing asynchronous stress-related patterns across modalities to be effectively captured.

\section{Methodology}

\subsection{Dataset Description}
We use the publicly available distracted driving dataset of Taamneh et al.~\cite{Distrdriving}, which contains synchronized facial videos, physiological signals, gaze measurements, and driving annotations collected in a simulator. We focus on Normal Driving (ND) as baseline and Sensorimotor Distraction (MD), where texting tasks induce stress during P2 and P4, while P1, P3, and P5 are non-stress phases.





\begin{figure}[t]
\centering
\includegraphics[width=1.0\textwidth]{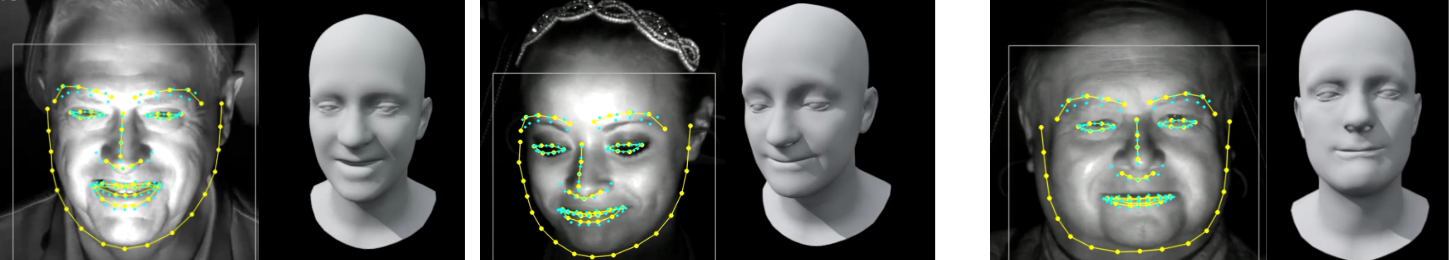}
\caption{\textit{3DMM-based feature visualization} Left: infrared frames with MediaPipe landmarks. Right: mean-face rendering driven by 3DMM expression and pose parameters, isolating facial motion from identity.}

\label{fig:featureviz}
\end{figure}


\subsection{Facial Feature Extraction}

\noindent\textbf{3D facial coefficients.}
We extract per-frame 3D facial expression and pose parameters from infrared facial videos using EMOCA~\cite{EMOCA:CVPR:2021,filntisis2022visual}, a SOTA monocular 3D face reconstruction framework built upon DECA~\cite{DECA:Siggraph2021}. For each frame $t$, EMOCA outputs a 56-dimensional vector
$\mathbf{x}_t = [\,\mathbf{exp}_t \in \mathbb{R}^{50},\ \mathbf{pose}_t \in \mathbb{R}^{6}\,]$,
where $\mathbf{exp}_t$ encodes facial expression coefficients and $\mathbf{pose}_t$ captures head and jaw pose. Following prior work~\cite{RingNet:CVPR:2019}, we retain only the first 50 expression coefficients, which capture the dominant expressive variation.


Figure~\ref{fig:featureviz} illustrates the extracted parameters projected onto a mean FLAME mesh with identity parameters set to zero and rendered alongside the input video frames with MediaPipe landmark overlays for qualitative inspection.

\noindent\textbf{Temporal derivatives.}
To capture dynamic facial behavior, we compute frame-to-frame differences of expression and pose parameters,
$\Delta \mathbf{exp}_t = \mathbf{exp}_t - \mathbf{exp}_{t-1}$ and
$\Delta \mathbf{pose}_t = \mathbf{pose}_t - \mathbf{pose}_{t-1}$,
which emphasize rapid motion changes and expressive transitions.

\subsection{Physiological Features and Gaze Dynamics}

To complement facial behavior, we use heart rate, breathing rate, and perinasal perspiration as a second modality.

When gaze measurements are available, given 2D gaze positions $(x_t, y_t)$, horizontal and vertical gaze velocities are computed by finite differences as
$v^{x}_t = (x_t - x_{t-1}) / \Delta t$ and
$v^{y}_t = (y_t - y_{t-1}) / \Delta t$,
with gaze speed defined as $\|v_t\| = \sqrt{(v^{x}_t)^2 + (v^{y}_t)^2}$. Gaze accelerations are computed analogously as
$a^{x}_t = (v^{x}_t - v^{x}_{t-1}) / \Delta t$ and
$a^{y}_t = (v^{y}_t - v^{y}_{t-1}) / \Delta t$,
with magnitude $\|a_t\| = \sqrt{(a^{x}_t)^2 + (a^{y}_t)^2}$. Short-term temporal statistics include rolling mean and standard deviation of gaze speed over $1$\,s and $3$\,s windows, as well as a $2$\,s gaze dispersion measure defined as $\|(\sigma_{x_t}, \sigma_{y_t})\|_2$.

\subsection{Statistical Analysis}
\label{StatisticalAnalysis}
We first conduct a statistical analysis to examine how facial motion relates to stress during driving. In particular, we study correlations between facial expression and pose parameters and stress conditions, using correlation measures consistent with those employed in prior physiological stress analyses \cite{pavlidis1}. 


\noindent\textbf{Phase-wise MD--ND effects.}
We quantify stress-related effects by paired comparisons between sensorimotor distracted (MD) and normal driving (ND) across five predefined phases. P2 and P4 correspond to stimulus intervals, P3 to a fixed detour segment (4.4--5.6,km, mapped to time), and P5 to a 120,s recovery period. The remaining intervals are treated as non-stress phases. For each subject, phase, and modality (breathing rate, heart rate, perinasal perspiration), phase means are computed after quality control and MD--ND differences $\Delta_{s,p,m}$ are assessed using two-sided one-sample $t$-tests against zero. The same phase-wise paired-difference protocol is applied to all facial expression and pose parameters.

\begin{figure}[t]
\centering
\includegraphics[width=1.0\textwidth]{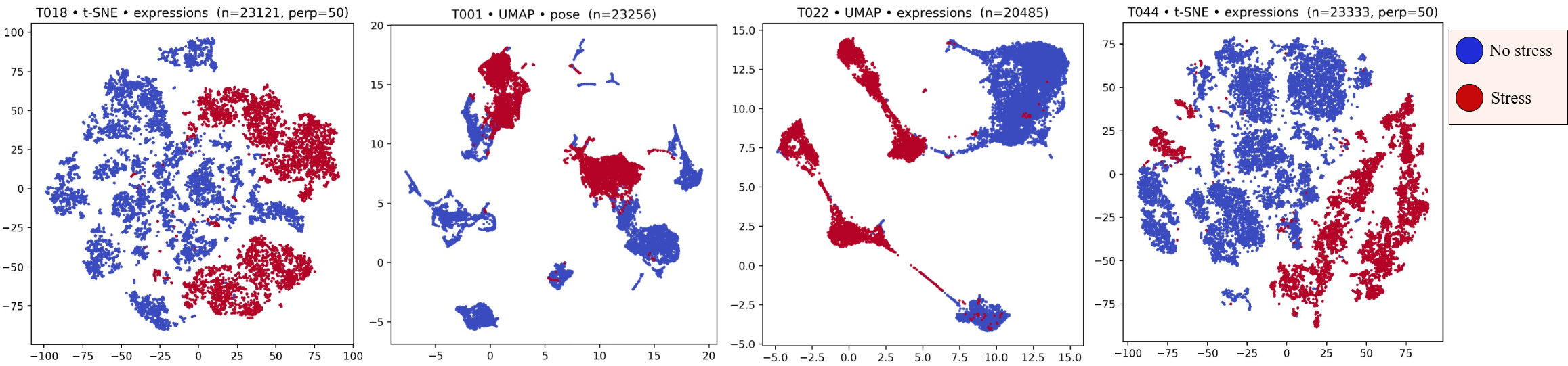}
\caption{From left to right: t-SNE and UMAP embeddings of 3D facial expression and pose parameters for subjects T018, 1001, T022, T044. Blue points denote non-stress and red points denote stress samples, revealing clear stress-related structure at the subject level.}
\label{fig:tsne}
\end{figure}

\noindent\textbf{Qualitative feature structure.}
To inspect the structure of the facial feature space, we apply t-SNE and UMAP to expression and pose parameters. As illustrated in Fig.~\ref{fig:tsne}, subject-wise embeddings reveal consistent separation between stress and non-stress samples relative to individual baselines.


\noindent\textbf{Temporal smoothing and facial dynamics.}
Facial expression and pose parameters are treated as temporal signals $X_{s,p}(t)$. For each parameter, we compute a mean level and a velocity summary  $\nu=\frac{1}{T}\sum_t |\dot{\tilde{Y}}(t)|$ from a smoothed signal $\tilde{Y}(t)$. We evaluate three temporal operators: no smoothing, symmetric triangular convolution, and cubic smoothing splines. Paired MD--ND differences $\Delta\mu_{s,p,f}$ and $\Delta\nu_{s,p,f}$ of mean and velocity summaries are tested across subjects using one-sample $t$-tests.

\begin{figure}[t]
\centering
\includegraphics[width=1.0\textwidth]{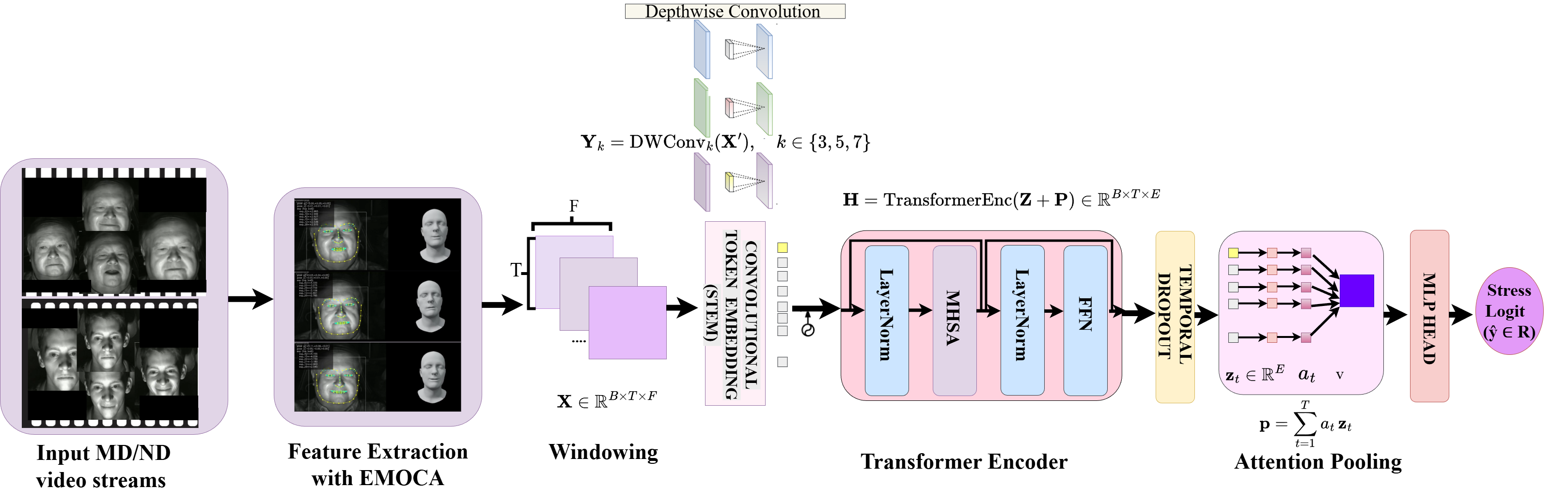}
\caption{Overview of the proposed visual stress recognition pipeline, using 3D facial features with first-order normalization, Transformer-based temporal modeling, and attention pooling for stress classification.}
\label{fig:conceptualframework}
\end{figure}

\subsection{Transformer Temporal Modeling}
\label{crossmodal}

While the preceding analysis uses classical statistical tools to study facial dynamics and stress, such methods are limited in modeling complex temporal patterns and multimodal interactions. Our approach follows the pipeline in Fig.~\ref{fig:conceptualframework}, combining Transformer encoders with attention pooling for stress prediction. Signals are temporally aligned, cleaned to ensure monotonic timestamps, and segmented into non-overlapping 9\,s windows, which outperformed shorter, longer, and overlapping alternatives in our validation experiments.


Let $\mathbf{x}^{d}_{s,p,w}(t)\in\mathbb{R}^{F_0}$ denote the per-frame 3D facial parameters for subject $s$, phase $p$, window $w$, and time index $t$, extracted from drive $d\in\{\mathrm{MD},\mathrm{ND}\}$. Each frame comprises expression coefficients $\mathbf{exp}(t)$ and pose parameters $\mathbf{pose}(t)$, and first-order temporal differences
$\Delta \mathbf{x}^{d}_{s,p,w}(t)=\mathbf{x}^{d}_{s,p,w}(t)-\mathbf{x}^{d}_{s,p,w}(t-1)$ for $t=2,\dots,T$, which capture facial motion dynamics.
Accordingly, the frame-level visual sequence consists of:
\begin{equation}
\mathbf{X}^{\mathrm{MD}}_{s,p,w}
=
\big[\mathbf{exp}^{\mathrm{MD}}_{s,p,w}(t)\;\|\;\mathbf{pose}^{\mathrm{MD}}_{s,p,w}(t)\;\|\;\Delta\mathbf{x}^{\mathrm{MD}}_{s,p,w}(t)\big]_{t=1}^{T}
\in\mathbb{R}^{T\times F}.
\end{equation}



To account for subject-specific baseline behavior, we use the paired ND drive to compute a window-level velocity-difference descriptor,
$\Delta\mathbf{v}{s,p,w}=\tfrac{1}{T-1}\sum{t=2}^{T}!\left(\Delta\mathbf{x}^{\mathrm{MD}}{s,p,w}(t)-\Delta\mathbf{x}^{\mathrm{ND}}{s,p,w}(t)\right)$.
This first-order normalization emphasizes stress-induced deviations from normal driving and is appended to the model input.

A multiscale convolutional stem extracts short-term temporal patterns using parallel depthwise 1D convolutions with kernel sizes $3$, $5$, and $7$, whose outputs are concatenated and linearly projected to an embedding of dimension $E$:
\begin{equation}
\mathbf{Z}_0=\mathrm{Conv}_{1\times1}\big([\mathrm{Conv}_3(\mathbf{X})\;\|\;\mathrm{Conv}_5(\mathbf{X})\;\|\;\mathrm{Conv}_7(\mathbf{X})]\big).
\end{equation}
Learned positional embeddings are added, and the resulting sequence is processed by a stack of Transformer encoder layers with multi-head self-attention.

A fixed-length window representation is obtained via attention-based temporal pooling, with weights $a_t=\mathrm{softmax}(\mathbf{v}^\top\tanh(W\mathbf{z}_t))$ and pooled embedding $\mathbf{h}=\sum_{t=1}^{T}a_t\,\mathbf{z}_t\in\mathbb{R}^E$, emphasizing stress-relevant temporal segments. The embedding is passed to a lightweight feed-forward head and trained using binary cross-entropy with logits.

\noindent\textbf{Early fusion.}
As a strong multimodal baseline, we implement an \emph{early-fusion} strategy where 3D expression and pose parameters are concatenated with the secondary modality (either physiological features or gaze-dynamics) at the input level to form a single per-frame feature vector; the same windowing and normalization protocol is applied.

\begin{figure}[t]
\centering
\includegraphics[width=1.0\textwidth]{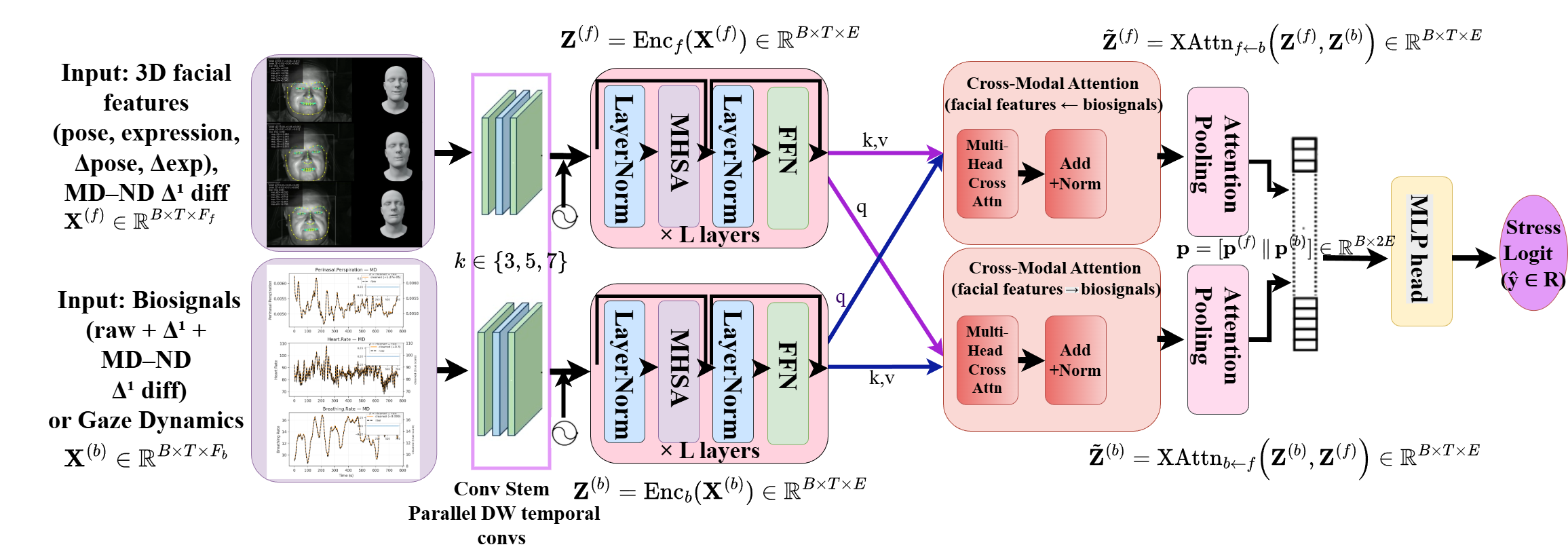}
\caption{Our proposed cross-modal attention fusion architecture, where 3D facial features and biosignal (or gaze-dynamics) streams are independently encoded with convolutional stems and Transformer encoders, fused via bidirectional cross-attention, aggregated using attention pooling, and finally concatenated for stress prediction.}
\label{fig:crossmodal}
\end{figure}

\noindent\textbf{Cross-modal attention fusion.}
To model inter-modal interactions, we employ a cross-modal attention architecture (Fig.~\ref{fig:crossmodal}) with two modality-specific encoders (facial features and the paired modality), each comprising a convolutional stem and a Transformer encoder. The latent sequences are fused via bidirectional cross-attention (facial features $\leftarrow$ modality-B and modality-B $\leftarrow$ facial features), followed by attention pooling per stream and concatenation of pooled representations for classification. 



\section{Experiments}

\subsection{Statistical Analysis: Physiological and Visual Stress Trackers}

\begin{figure}[!t]
    \centering \includegraphics[width=\linewidth,height=0.78\textheight,keepaspectratio]{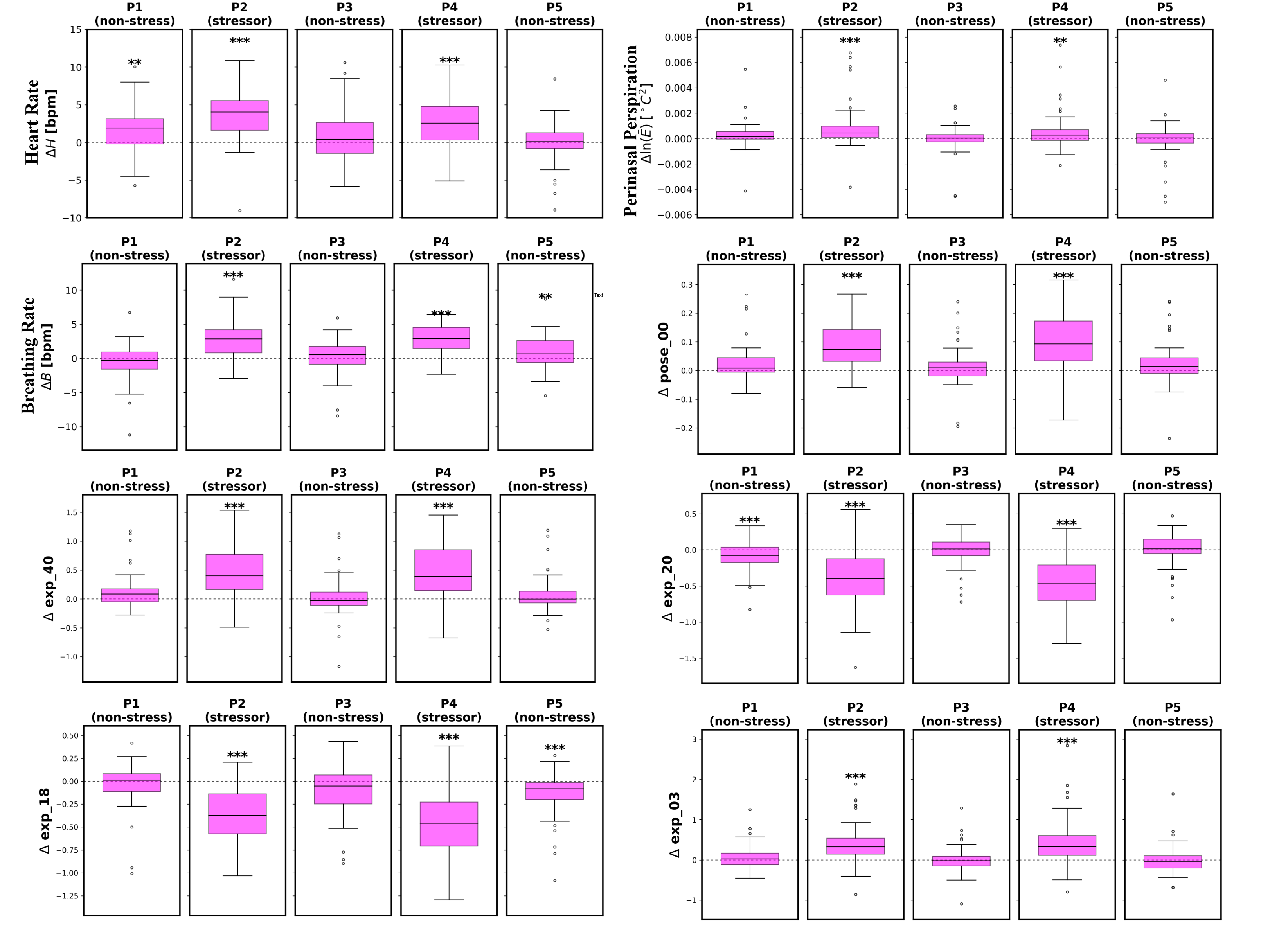}
    \caption{Phase-wise MD--ND differences illustrating stress-related separability in phases P2 and P4. Established physiological markers (heart rate, perinasal perspiration) and selected 3D facial parameters (pose\_00, exp\_40, exp\_20, exp\_18, exp\_03), identified via PCA for their strong correlation with stress, show pronounced deviations from baseline.}
    \label{fig:pairedttests}
\end{figure}

\noindent\textbf{Visual stress trackers (facial features coefficients).}
We applied the MD--ND differencing strategy (Section~\ref{StatisticalAnalysis}) to the 3D facial features and evaluated paired effects using two-sided one-sample $t$-tests. Figure~\ref{fig:pairedttests} summarizes the phase-wise results. At $p<0.001$, 24/50 expression coefficients show significant modulation in at least one stressor phase (P2 or P4), including 18/50 significant in both, while 2/6 pose parameters also exhibit stress-related effects. At $p<0.05$, 28/50 expression coefficients are significant in both stressor phases with an additional 3/50 significant in either P2 or P4, and 3/6 pose parameters show consistent modulation. Established physiological markers (perinasal perspiration, heart rate, and breathing rate) exhibit structured stress responses (Table~\ref{tab:comparisonstatistics}).

\begin{figure}[t]
\centering
\includegraphics[width=0.7\textwidth]{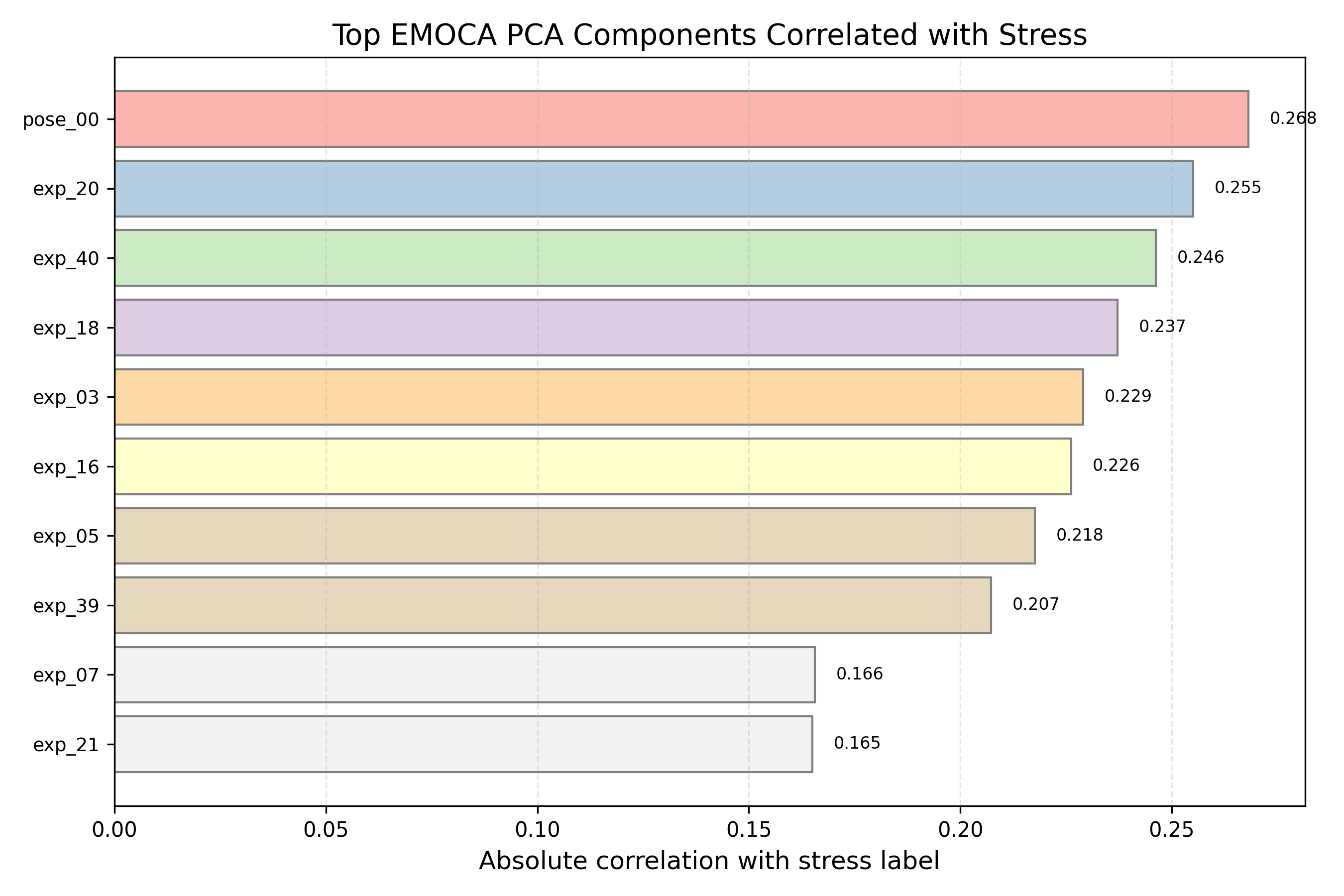}
\caption{Top facial features PCA components most correlated with stress.The strongest effects are observed for the global pose component \texttt{pose\_00} and the expression component \texttt{exp\_20}.}
\label{fig:pca_top10}
\end{figure}

\begin{figure}[t]
\centering
\includegraphics[width=0.72\textwidth]{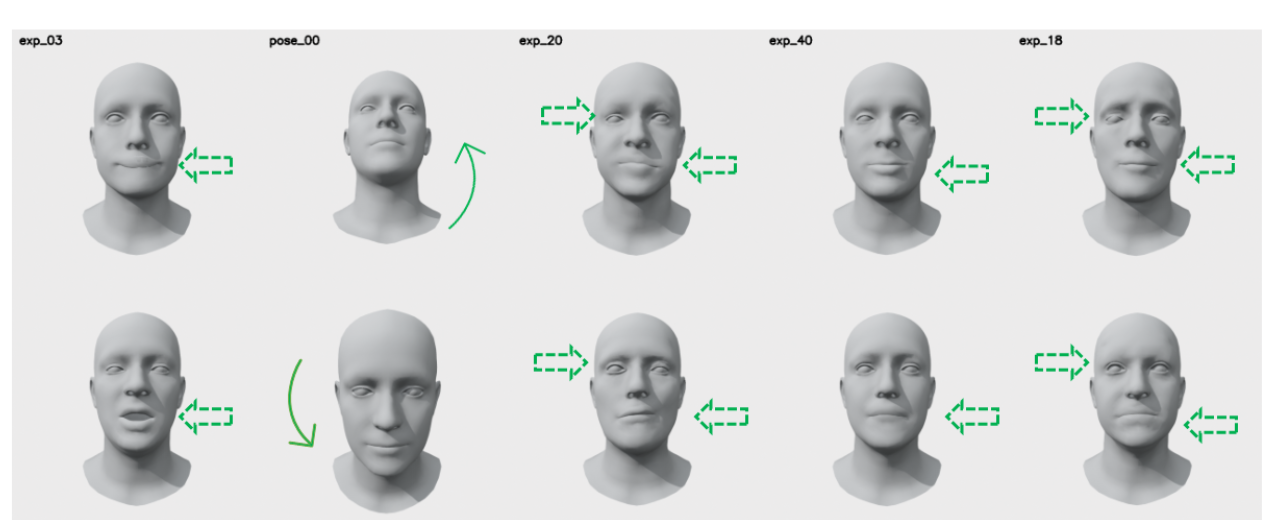}
\caption{Top stress-correlated components of the 3D facial behavior feature vector visualized on the mean face. The mean face is perturbed along the direction of each selected component to illustrate its geometric effect. Green arrows highlight the dominant facial regions and directions of deformation (primarily mouth, eyes, and head rotation).
}
\label{fig:pose_exp_vis}
\end{figure}

To examine the relationship between facial parameters and stress, we performed PCA on the expression and pose parameters. This analysis highlighted \texttt{pose\_00} and \texttt{exp\_40} as the dimensions most strongly correlated with stress (Fig.~\ref{fig:pca_top10}). The five most stress-related components are visualized by perturbing the mean face along the corresponding principal directions (Fig.~\ref{fig:pose_exp_vis}), illustrating the facial deformations associated with stress.


\begin{table}[t]
\centering
\caption{
Phase-wise MD--ND significance ($p$-values) for physiological signals and PCA-selected facial parameters, showing pronounced effects in stressor phases (P2, P4).}
\setlength{\tabcolsep}{8pt}
\renewcommand{\arraystretch}{1.15}
\resizebox{\columnwidth}{!}
{
\begin{tabular}{l
                S[table-format=1.2e-2]
                S[table-format=1.2e-2]
                S[table-format=1.2e-1]
                S[table-format=1.2e-2]
                S[table-format=1.4]}
\hline
\textbf{Signal}
& \textbf{P1}
& \textbf{P2 (Stressor)}
& \textbf{P3}
& \textbf{P4 (Stressor)}
& \textbf{P5} \\
\hline

Breathing Rate
& 0.1668
& \textbf{2.02e-12}
& 0.4036
& \textbf{5.36e-15}
& \textbf{0.00744} \\

Heart Rate
& \textbf{0.00107}
& \textbf{1.75e-10}
& 0.2015
& \textbf{3.34e-08}
& 0.6529 \\

Perinasal Perspiration
& 0.08623
& \textbf{2.99e-04}
& 0.7907
& \textbf{0.005998}
& 0.5452 \\

\hline
Expression03 (\texttt{exp\_03})
& 0.0798
& \textbf{5.72e-08}
& 0.79
& \textbf{4.66e-07}
& 0.905 \\

Expression18 (\texttt{exp\_18})
& 0.2056
& \textbf{7.85e-13}
& 0.0132
& \textbf{1.25e-13}
& \textbf{0.00037} \\

Expression20 (\texttt{exp\_20})
& \textbf{0.00087}
& \textbf{4.29e-11}
& 0.5105
& \textbf{1.00e-13}
& 0.9818 \\

Expression40 (\texttt{exp\_40})
& 0.00105
& \textbf{4.48e-12}
& 0.7244
& \textbf{3.05e-10}
& 0.0674 \\

Pose0 (\texttt{pose\_00})
& 0.00405
& \textbf{1.23e-10}
& 0.1335
& \textbf{8.34e-11}
& 0.0223 \\
\hline

\label{tab:comparisonstatistics}
\end{tabular}
}

\end{table}

\begin{figure}[t]
    \centering
    \includegraphics[width=1.0\textwidth]{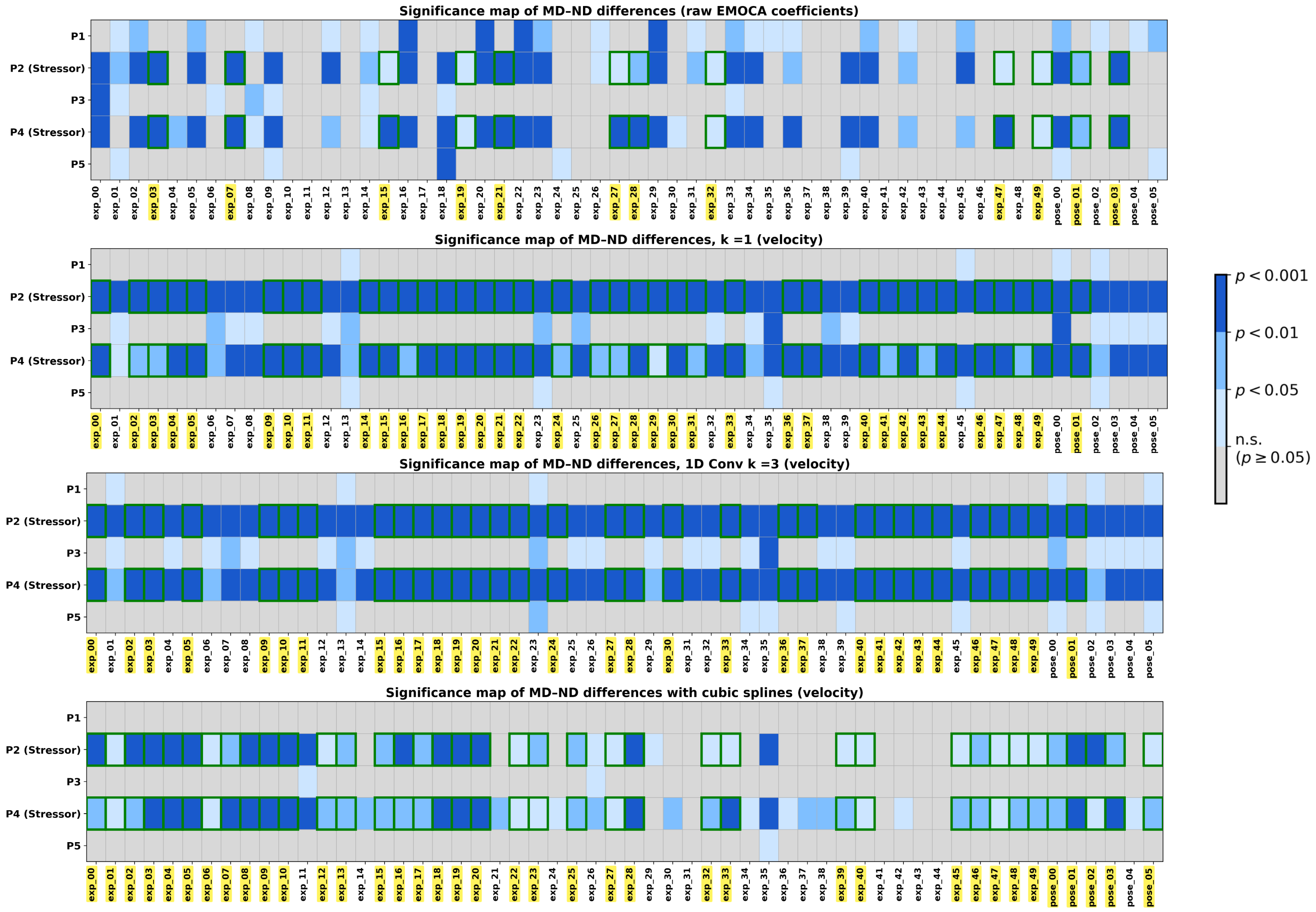}
    \caption{Phase-wise MD--ND significance maps for 3D facial features under different temporal operators. Without smoothing ($k=1$), 37 features show strong stressor-phase significance (predominantly $p<0.001$), whereas cubic spline smoothing yields a similar number of significant coefficients (38) but with higher $p$-values, indicating more conservative effects.
}

    \label{fig:emoca_conv}
\end{figure}

\noindent\textbf{1D Convolution and Spline Embeddings.}
Figure~\ref{fig:emoca_conv} compares phase-wise MD--ND significance patterns for 3D facial expression and pose features under different temporal operators. Mean-level features show weak effects, whereas velocity features strongly discriminate the stressor phases (P2, P4). Without smoothing ($k{=}1$), all 56 features are significant in both stressor phases, while cubic spline smoothing yields more conservative patterns, with 38 expression and pose parameters remaining significant in both P2 and P4. Overall, velocity-based representations are most informative, indicating that stress is primarily reflected in rapid frame-to-frame facial dynamics. 


\subsection{LDA Stress Axis and Geometric Visualization}

\begin{figure}[t]
\centering
\includegraphics[width=0.5\textwidth]{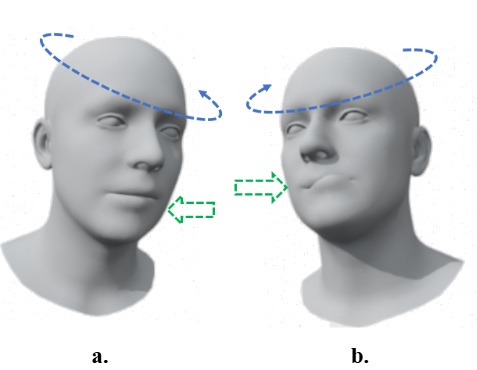}
\caption{Visualization of the LDA stress discriminant direction in EMOCA space. The expression coefficients and pose parameters are perturbed by $-3\sigma$ (a) and $+3\sigma$ (b) along the LDA axis and rendered on the mean FLAME face.
}
\label{fig:LDAtriptych}
\end{figure}

To identify which facial components are most predictive of stress, we applied binary  Linear Discriminant Analysis (LDA) to the standardized coefficients.  Because the problem is two-class, LDA yields a single discriminant direction $w$ that maximally separates stress from no-stress frames, producing for each sample  a scalar projection $z_i = w^\top x^{\text{std}}_i$, which effectively serves  as a latent action unit for stress. To visualize the effect of this axis on  3D geometry, we perturbed the mean FLAME face by moving the facial features by $\pm 3\sigma$ along $w$ and rendered the resulting shapes. As shown in Fig.~\ref{fig:LDAtriptych} 
the LDA stress axis is primarily associated with lower-face deformations (mouth, jaw) and head pose variation, indicating that these components contribute most strongly to discriminating stress from non-stress states.

\subsection{Classification Metrics}

All reported results were obtained using 5-fold subject-wise cross-validation to ensure no data leakage between sets. Each window was assigned a binary stress label based on the proportion of stressed frames it contained: windows with a stress ratio greater than 0.4 were labeled as stress. Binary classification follows the phase-based protocol, whereas continuous prediction would require finer-grained annotations. Performance was computed at the window level, and metrics including AUROC, AUPRC, F1, Accuracy, and Balanced Accuracy were averaged across folds. For all experiments, subject-wise normalization is applied per cross-validation split: per-feature statistics are estimated exclusively from the training windows of each split and used to normalize the corresponding training, validation, and test data, preventing information leakage across subjects.


\noindent\textbf{Early-Fusion Extensions.} We optimized temporal and training hyperparameters and obtained our strongest visual-only performance using non-overlapping 9\,s windows, dropout $=0.2$, and early stopping within 20 epochs. This configuration achieved a mean AUROC of $0.908 \pm 0.015$ and accuracy of $0.841 \pm 0.017$ under 5-fold subject-wise cross-validation (Table~\ref{tab:single_modality_5fold}). The visual representation includes expression and pose parameters, their first-order temporal differences, and first order normalization computed from these dynamics, enabling the model to capture stress-induced deviations relative to baseline driving.

Early fusion does not outperform the visual-only baseline (AUROC $0.901 \pm 0.017$ for facial features+Bio), indicating that facial dynamics already capture the dominant stress-related information under this setting.

\noindent\textbf{Comparisons.}
To ensure a fair comparison with established stress-classification approaches, we evaluated our model against two literature-aligned baselines under the same 5-fold subject-wise protocol. First, we implemented \emph{StressID-style} traditional ML pipelines using non-overlapping windows and per-window mean and standard deviation features, training SVM and kNN classifiers with subject-wise cross-validation. Second, we evaluated a fully-connected baseline described by Giannakakis et al \cite{rel27} adapted for binary stress prediction, using identical normalization and splits.

\begin{table}[t]
\centering
\caption{Comparison of stress classification performance between the proposed method and prior baselines. Results are reported as mean $\pm$ standard deviation over 5-fold subject-wise cross-validation. BIO denotes physiological features and EF denotes early fusion.}
\label{tab:comparison_ours_vs_baselines}
\setlength{\tabcolsep}{2.5pt}
\renewcommand{\arraystretch}{1.05}
\resizebox{\columnwidth}{!}
{
\begin{tabular}{lccccc}
\hline
 \textbf{Modality}
& \textbf{AUROC}
& \textbf{AUPRC}
& \textbf{F1}
& \textbf{Accuracy}
& \textbf{Balanced Acc.} \\
\hline

\multicolumn{6}{c}{\textit{Ours}} \\
\hline

facial features
& \textbf{0.908} $\pm$ 0.015
& 0.915 $\pm$ 0.022
& \textbf{0.848} $\pm$ 0.025
& \textbf{0.8412} $\pm$ 0.0168
& \textbf{0.8407} $\pm$ 0.0183 \\

facial features + Bio EF 
& $0.901 \pm 0.017$
& $\textbf{0.918} \pm 0.024$
& $0.832 \pm 0.039$
& $0.831 \pm 0.036$
& $0.835 \pm 0.033$ \\

\hline
\multicolumn{6}{c}{\textit{StressID reported traditional ML baselines (9s windows; mean+std statistics) \cite{NEURIPS2023_5f09bfe6}}} \\
\hline

facial features (SVM)
& $0.893 \pm 0.020$
& $0.903 \pm 0.026$
& $0.842 \pm 0.029$
& $0.829 \pm 0.027$
& $0.826 \pm 0.026$ \\

facial features (kNN)
& $0.848 \pm 0.037$
& $0.843 \pm 0.050$
& $0.803 \pm 0.0366$
& $0.790 \pm 0.0372$
& $0.789 \pm 0.0373$ \\

facial features + Bio EF (SVM)
& $0.893 \pm 0.021$
& $0.903 \pm 0.026$
& $0.840 \pm 0.027$
& $0.827 \pm 0.024$
& $0.824 \pm 0.022$ \\

facial features + Bio EF (kNN)
& $0.856 \pm 0.033$
& $0.849 \pm 0.042$
& $0.812 \pm 0.025$
& $0.797 \pm 0.027$
& $0.795 \pm 0.027$ \\

\hline
\multicolumn{6}{c}{\textit{Fully-connected baselines (Giannakakis et al. MLP \cite{rel27})}} \\
\hline
facial features
& $0.871 \pm 0.023$
& $0.889 \pm 0.032$
& $0.822 \pm 0.031$
& $0.810 \pm 0.028$
& $0.809 \pm 0.026$ \\

facial features + Bio EF 
& $0.8757 \pm 0.0206$
& $0.8921 \pm 0.0276$
& $0.8148 \pm 0.0352$
& $0.8051 \pm 0.0280$
& $0.8046 \pm 0.0252$ \\

\hline
\end{tabular}
}
\end{table}

As shown in Table~\ref{tab:comparison_ours_vs_baselines}, traditional SVM-based baselines are competitive, with facial features (SVM) achieving AUROC $0.893 \pm 0.020$ and early-fusion facial features+Bio (SVM) reaching AUROC $0.893 \pm 0.021$. Our visual-only Transformer slightly improves upon these results (AUROC $0.908 \pm 0.015$), while maintaining higher accuracy and balanced accuracy. In contrast, fully-connected baselines perform consistently worse under the same subject-wise protocol.

\noindent\textbf{Cross-Modal Attention Fusion.} We evaluate our cross-modal attention architecture \ref{crossmodal} that processes visual and non-visual signals with modality-specific temporal encoders, followed by bidirectional cross-attention. This design enables each modality to attend selectively to temporally relevant cues in the other before aggregation via attention pooling and final classification.

Cross-modal fusion consistently outperforms both single-modality and early-fusion baselines when using the full set of visual facial descriptors and all available physiological signals, combined with subject-wise normalization. As shown in Table~\ref{tab:single_modality_5fold}, facial features combined with physiological signals via cross-attention achieves the strongest overall performance, with AUROC $0.92 \pm 0.04$, F1 $0.866 \pm 0.054$, accuracy $0.866 \pm 0.05$, and balanced accuracy $0.87 \pm 0.05$. Although the biosignal-only branch performs weakly in isolation, physiological cues remain complementary and become useful when their temporal interaction with facial dynamics is modeled explicitly through cross-modal. Cross-modal fusion with gaze yields competitive but slightly lower performance (AUROC $0.918 \pm 0.043$), while combinations excluding facial dynamics perform substantially worse.

\begin{table}[t]
\centering
\caption{Stress classification performance using single-modality, early-fusion, and cross-modal attention inputs.  Evaluation is performed with 5-fold subject-wise cross-validation. Results are reported as mean $\pm$ standard deviation.}
\label{tab:single_modality_5fold}
\resizebox{\columnwidth}{!}
{
\begin{tabular}{lccccc}
\hline
\textbf{Modality} 
& \textbf{AUROC} 
& \textbf{AUPRC} 
& \textbf{F1} 
& \textbf{Accuracy} 
& \textbf{Balanced Acc.} \\
\hline
facial features
& 0.908 $\pm$ 0.015
& 0.915 $\pm$ 0.022
& 0.848 $\pm$ 0.025
& 0.8412 $\pm$ 0.0168
& 0.8407 $\pm$ 0.0183 \\

Bio (PP, HR, BR)
& 0.527 $\pm$ 0.054
& 0.568 $\pm$ 0.023
& 0.439 $\pm$ 0.324
& 0.510 $\pm$ 0.041
& 0.498 $\pm$ 0.003 \\

Early Fusion(facial features + Bio)
& $0.901 \pm 0.017$
& $0.918 \pm 0.024$
& $0.832 \pm 0.039$
& $0.831 \pm 0.036$
& $0.835 \pm 0.033$ \\

Early Fusion(facial features + Gaze)
& 0.863 $\pm$ 0.047
& 0.866 $\pm$ 0.074
& 0.780 $\pm$ 0.067
& 0.780 $\pm$ 0.052
& 0.785 $\pm$ 0.048 \\

\hline
Cross-Modal (facial features + Bio)
& \textbf{0.92} $\pm$ \textbf{0.04}
& 0.91 $\pm$ 0.05
& \textbf{0.866} $\pm$ \textbf{0.054}
& \textbf{ 0.866} $\pm$ \textbf{0.05}
& \textbf{0.87} $\pm$ \textbf{0.05} \\

Cross-Modal (facial features + Gaze)
& 0.918 $\pm$ 0.043
& \textbf{0.92} $\pm$ \textbf{0.04}
& 0.85 $\pm$ 0.05
& 0.855 $\pm$ 0.047
& 0.857 $\pm$ 0.046 \\

Cross-Modal (Gaze + Bio)
& 0.65 $\pm$ 0.039
& 0.64 $\pm$ 0.06
& 0.55 $\pm$ 0.08
& 0.591 $\pm$ 0.03
& 0.59 $\pm$ 0.03 \\
\hline
\end{tabular}
}
\end{table}

\section{Conclusions}

This work studies stress estimation under distracted driving by jointly analyzing disentangled 3D facial features, physiological signals, and gaze dynamics. Phase-wise analysis shows that several facial expression and pose parameters exhibit stress-specific modulations comparable to physiological markers, with stress encoded mainly in temporal dynamics, particularly velocity-based descriptors. Motivated by this, we evaluate unimodal, early-fusion, and cross-modal Transformer models, showing that multimodal fusion performs best, with facial--physiological cross-modal attention reaching AUROC $0.92 \pm 0.04$ and accuracy $0.866 \pm 0.05$. Visual-only models remain competitive when physiological signals are unavailable. Limitations include evaluation on a simulated driving dataset; future work will validate naturalistic settings, improve interpretability, explore end-to-end temporal learning, and investigate real-time deployment via causal windowing and lightweight 3D reconstruction pipelines.





\subsubsection{Acknowledgements} 
The authors thank Ioannis Pavlidis for valuable discussions on the Distracted Driving dataset \cite{Distrdriving} and relevant biosignal analysis. This work was co-funded by the 2025 FORTH Synergy Grant "AI SPACE HERITAGE" and the VMware University Research Fund (VMURF). 

%
%
%
\bibliographystyle{splncs04}
\bibliography{bib/paper}

\begin{thebibliography}{10}
\providecommand{\url}[1]{\texttt{#1}}
\providecommand{\urlprefix}{URL }
\providecommand{\doi}[1]{https://doi.org/#1}

\bibitem{rel33}
Almeida, J., Rodrigues, F.: Facial expression recognition system for stress detection with deep learning. pp. 256--263 (01 2021)

\bibitem{Bustos}
Bustos, C., Sole-Ribalta, A., Elhaouij, N., Borge-Holthoefer, J., Lapedriza, A., Picard, R.: Analyzing the visual road scene for driver stress estimation. IEEE Transactions on Affective Computing  \textbf{16}(3),  1787--1801 (2025). \doi{10.1109/TAFFC.2025.3539003}

\bibitem{NEURIPS2023_5f09bfe6}
Chaptoukaev, H., Strizhkova, V., Panariello, M., Dalpaos, B., Reka, A., Manera, V., Th\"{u}mmler, S., Ismailova, E., W., N., bremond, f., Todisco, M., Zuluaga, M.A., M.~Ferrari, L.: Stressid: a multimodal dataset for stress identification. In: NeurIPS. vol.~36, pp. 29798--29811 (2023)

\bibitem{EMOCA:CVPR:2021}
Danecek, R., Black, M.J., Bolkart, T.: {EMOCA}: {E}motion driven monocular face capture and animation. In: CVPR. pp. 20311--20322 (2022)

\bibitem{rel32}
Ding, D., Xu, W., Liu, X., Zhu, T.: Facial video based stress detection for enhancing ecological validity. Acta Psychologica  \textbf{255} (2025)

\bibitem{DECA:Siggraph2021}
Feng, Y., Feng, H., Black, M.J., Bolkart, T.: Learning an animatable detailed {3D} face model from in-the-wild images. SIGGRAPH  \textbf{40}(8) (2021)

\bibitem{filntisis2022visual}
Filntisis, P.P., Retsinas, G., Paraperas-Papantoniou, F., Katsamanis, A., Roussos, A., Maragos, P.: Visual speech-aware perceptual 3d facial expression reconstruction from videos. arXiv:2207.11094  (2022)

\bibitem{rel25}
Gavrilescu, M., Vizireanu, N.: Predicting depression, anxiety, and stress levels from videos using the facial action coding system. Sensors  \textbf{19}(17) (2019)

\bibitem{rel31}
Giannakakis, G., Pediaditis, M., Manousos, D., Kazantzaki, E., Chiarugi, F., Simos, P., Marias, K., Tsiknakis, M.: Stress and anxiety detection using facial cues from videos. Biomedical Signal Processing and Control  \textbf{31},  89--101 (2017)

\bibitem{Giannakakis2019ReviewOP}
Giannakakis, G., Grigoriadis, D., Giannakaki, K., Simantiraki, O., Roniotis, A., Tsiknakis, M.: Review on psychological stress detection using biosignals. IEEE Trans. on Affective Computing  \textbf{13},  440--460 (2019)

\bibitem{rel29}
Giannakakis, G., Koujan, M.R., Roussos, A., Marias, K.: Automatic stress detection evaluating models of facial action units. In: FG. p. 728–733 (2020)

\bibitem{rel27}
Giannakakis, G., Koujan, M.R., Roussos, A., Marias, K.: Automatic stress analysis from facial videos based on deep facial action units recognition. Pattern Anal. Appl. p. 521–535 (Aug 2022)

\bibitem{Giannakakis2018EvaluationOH}
Giannakakis, G.A., Manousos, D., Chaniotakis, V., Tsiknakis, M.: Evaluation of head pose features for stress detection and classification. BHI pp. 406--409 (2018)

\bibitem{HasanCardiovascular2023}
Hasan, M.T., Alghamdi, H., Taamneh, S., Manser, M., Wunderlich, R., Tsiamyrtzis, P., Pavlidis, I.: Investigating cardiovascular activation of young adults in routine driving. IEEE Transactions on Affective Computing  \textbf{15}(3),  769--786 (2024). \doi{10.1109/TAFFC.2023.3291330}

\bibitem{ecsa-7-08227}
Hazer-Rau, D., Zhang, L., Traue, H.C.: A workflow for affective computing and stress recognition from biosignals. Engineering Proceedings  \textbf{2}(1) (2020)

\bibitem{PHYSIOLOGICAL}
Hota, A., Park, S.W.: Stress detection using physiological signals based on machine learning. In: 2022 International Conference on Computational Science and Computational Intelligence (CSCI). pp. 379--384 (2022). \doi{10.1109/CSCI58124.2022.00074}

\bibitem{Huynhetal}
Huynh, T., Manser, M., Pavlidis, I.: Arousal responses to regular acceleration events divide drivers into high and low groups: A naturalistic pilot study of accelarousal and its implications to human-centered design. In: Extended Abstracts of the 2021 CHI Conference on Human Factors in Computing Systems. CHI EA '21, Association for Computing Machinery, New York, NY, USA (2021). \doi{10.1145/3411763.3451809}, \url{https://doi.org/10.1145/3411763.3451809}

\bibitem{muse}
Jaiswal, M., Bara, C.P., Luo, Y., Burzo, M., Mihalcea, R., Provost, E.M.: Muse: a multimodal dataset of stressed emotion. In: Int'l Conf. on Language Resources and Evaluation (2020)

\bibitem{rel26}
Jaiswal, S., Valstar, M.: Deep learning the dynamic appearance and shape of facial action units. In: WACV. pp.~1--8 (2016)

\bibitem{rel23}
Jeon, T., Bae, H.B., Lee, Y., Jang, S., Lee, S.: Deep-learning-based stress recognition with spatial-temporal facial information. Sensors  \textbf{21}(22) (2021)

\bibitem{swellkw}
Koldijk, S., Sappelli, M., Verberne, S., Neerincx, M.A., Kraaij, W.: The swell knowledge work dataset for stress and user modeling research. In: ICMI (2014)

\bibitem{rel30}
Koujan, M.R., Alharbawee, L., Giannakakis, G., Pugeault, N., Roussos, A.: Real-time facial expression recognition “in the wild” by disentangling 3d expression from identity. In: FG. p. 24–31 (2020)

\bibitem{COMPLEMENTARY3}
Kumar, A., Karthik, G.M.: Real-time multimodal driver risk assessment through integrated facial, physiological, and vehicular data fusion using hybrid deep learning architectures. In: 2025 Third International Conference on Networks, Multimedia and Information Technology (NMITCON). pp.~1--7 (2025). \doi{10.1109/NMITCON65824.2025.11187444}

\bibitem{GLMDriveNet}
Liu, W., Gong, Y., Zhang, G., Lu, J., Zhou, Y., Liao, J.: Glmdrivenet: Global–local multimodal fusion driving behavior classification network. Eng. Appl. AI  (2024)

\bibitem{FMDNet}
Liu, W., Lu, J., Liao, J., Qiao, Y., Zhang, G., Zhu, J., Xu, B., Li, Z.: Fmdnet: Feature-attention-embedding-based multimodal-fusion driving-behavior-classification network. IEEE Trans. on Comp. Social Systems  \textbf{11}(5) (2024)

\bibitem{clas}
Markova, V., Ganchev, T., Kalinkov, K.: Clas: A database for cognitive load, affect and stress recognition (01 2020)

\bibitem{MOU2023121066}
Mou, L., Chang, J., Zhou, C., Zhao, Y., Ma, N., Yin, B., Jain, R., Gao, W.: Multimodal driver distraction detection using dual-channel network of cnn and transformer. Expert Systems with Applications  \textbf{234},  121066 (2023)

\bibitem{COMPLEMENTARY2}
Noh, B., Park, M., Han, Y., Kim, J.: A multi-modal approach for detecting drivers’ distraction using bio-signal and vision sensor fusion in driver monitoring systems. Engineering Applications of Artificial Intelligence  \textbf{161},  112265 (2025). \doi{https://doi.org/10.1016/j.engappai.2025.112265}, \url{https://www.sciencedirect.com/science/article/pii/S0952197625022730}

\bibitem{pavlidis1}
Pavlidis, I., Dcosta, M., Taamneh, S., Manser, M., Ferris, T., Wunderlich, R., Akleman, E., Tsiamyrtzis, P.: Dissecting driver behaviors under cognitive, emotional, sensorimotor, and mixed stressors. Scientific Reports  \textbf{6},  25651 (05 2016)

\bibitem{UBCuphys}
Sabour, R.M., Benezeth, Y., De~Oliveira, P., Chappé, J., Yang, F.: Ubfc-phys: A multimodal database for psychophysiological studies of social stress. IEEE Trans. on Affective Computing  \textbf{14}(1),  622--636 (2023)

\bibitem{RingNet:CVPR:2019}
Sanyal, S., Bolkart, T., Feng, H., Black, M.: Learning to regress 3d face shape and expression from an image without 3d supervision. In: CVPR (Jun 2019)

\bibitem{wesad}
Schmidt, P., Reiss, A., Duerichen, R., Marberger, C., Van~Laerhoven, K.: Introducing wesad, a multimodal dataset for wearable stress and affect detection. In: ICMI. p. 400–408 (2018)

\bibitem{Siam2023AutomaticSD}
Siam, A.I., Gamel, S.A., Talaat, F.M.: Automatic stress detection in car drivers based on non-invasive physiological signals using machine learning techniques. Neural Computing and Applications  \textbf{35},  12891--12904 (2023)

\bibitem{healthcare}
Sinhal, A., Sinhal, A., Sinhal, A.: Stress monitoring in healthcare: An ensemble machine learning framework using wearable sensor data (2025)

\bibitem{sus}
Steeneken, H.J.M., Hansen, J.H.L.: Speech under stress conditions: overview of the effect on speech production and on system performance. ICASSP  \textbf{4} (1999)

\bibitem{Distrdriving}
Taamneh, S., Tsiamyrtzis, P., Dcosta, M., Buddharaju, P., Khatri, A., Manser, M., Ferris, T., Wunderlich, R., Pavlidis, I.: A multimodal dataset for various forms of distracted driving. Scientific Data  \textbf{4},  170110 (08 2017)

\bibitem{TAVAKOLI2023101649}
Tavakoli, A., Lai, N., Balali, V., Heydarian, A.: How are drivers’ stress levels and emotions associated with the driving context? a naturalistic study. Journal of Transport and Health  \textbf{31},  101649 (2023). \doi{https://doi.org/10.1016/j.jth.2023.101649}, \url{https://www.sciencedirect.com/science/article/pii/S2214140523000853}

\bibitem{sadvaw}
Tran, T.D., Kim, J., Ho, N.H., Yang, H.J., Pant, S., Kim, S.H., Lee, G.S.: Stress analysis with dimensions of valence and arousal in the wild. Applied Sciences  \textbf{11}(11) (2021)

\bibitem{rel28}
Viegas, C., Lau, S.H., Maxion, R., Hauptmann, A.: Towards independent stress detection: A dependent model using facial action units. In: CBMI. pp.~1--6 (2018)

\bibitem{rel22}
Wang, X., Zhang, T., Chen, C.: Pau-net: Privileged action unit network for facial expression recognition. IEEE Trans. on Cognitive and Developmental Systems  \textbf{PP}, ~1--1 (01 2022)

\bibitem{COMPLEMENTARY1}
Widayat, T.A., Mintje, Q.A.P., Yosepha, S.Y.: Enhancing driver stress detection through multimodal integration of eye tracking and physiological signals. Logistica : Journal of Logistic and Transportation  \textbf{3}(3),  150–160 (Jul 2025). \doi{10.61978/logistica.v3i3.1147}, \url{https://journal.idscipub.com/index.php/logistica/article/view/1147}

\bibitem{MTASR}
Xu, J., Song, C., Yue, Z., Ding, S.: Facial video-based non-contact stress recognition utilizing multi-task learning with peak attention. IEEE Journal of Biomedical and Health Informatics  \textbf{28}(9),  5335--5346 (2024)

\bibitem{rel24}
Zhang, H., Feng, L., Li, N., Jin, Z., Cao, L.: Video-based stress detection through deep learning. Sensors  \textbf{20}(19) (2020)

\end{thebibliography}
%




\end{document}